\documentclass{article}

\usepackage{microtype}
\usepackage{graphicx}
\usepackage{subfigure}
\usepackage{booktabs} 
\usepackage{etoc}

\usepackage[utf8]{inputenc} 
\usepackage[T1]{fontenc}    
\usepackage{hyperref}       
\usepackage{url}            
\usepackage{booktabs}       
\usepackage{amsfonts}       
\usepackage{nicefrac}       
\usepackage{microtype}      

\usepackage{graphicx}
\usepackage{stfloats}
\usepackage{subfigure}
\usepackage{multirow}
\usepackage{amsmath}
\usepackage{float}
\usepackage{amsthm}
\usepackage{comment}
\usepackage{multicol}
\usepackage{appendix}
\usepackage{amssymb}
\usepackage{wrapfig}
\usepackage{bbm}
\usepackage{bbding}
\usepackage{framed}
\usepackage{setspace}
\usepackage{pifont}
\newcommand{\cmark}{\ding{51}}
\newcommand{\xmark}{\ding{55}}
\usepackage{makecell}

\usepackage[linesnumbered,ruled,lined]{algorithm2e}
\usepackage{textcomp,booktabs}

\newtheorem{theorem}{Theorem}

\newtheorem{assumption}{Assumption}

\usepackage{xcolor}         
\definecolor{mycolor1}{rgb}{0.82,0.70,0.54}
\definecolor{mycolor2}{rgb}{0.0,0.51,0.22}
\definecolor{mycolor3}{rgb}{0.80, 0.48, 0.37}
\definecolor{mycolor4}{rgb}{0.02, 0.33, 0.68}

\usepackage{hyperref}

\usepackage[accepted]{icml2023}

\usepackage{amsmath}
\usepackage{amssymb}
\usepackage{mathtools}
\usepackage{amsthm}

\usepackage[capitalize,noabbrev]{cleveref}

\theoremstyle{plain}

\theoremstyle{definition}
\theoremstyle{remark}

\usepackage[textsize=tiny]{todonotes}

\icmltitlerunning{Provable Dynamic Fusion for Low-Quality Multimodal Data}

\begin{document}

\twocolumn[
\icmltitle{Provable Dynamic Fusion for Low-Quality Multimodal Data}



\icmlsetsymbol{equal}{*}

\begin{icmlauthorlist}
\icmlauthor{Qingyang Zhang}{yyy}
\icmlauthor{Haitao Wu}{yyy}
\icmlauthor{Changqing Zhang}{yyy,lab}
\icmlauthor{Qinghua Hu}{yyy,lab}

\icmlauthor{Huazhu Fu}{comp1}
\icmlauthor{Joey Tianyi Zhou}{comp1,comp2}
\icmlauthor{Xi Peng}{sch}
\end{icmlauthorlist}

\icmlaffiliation{yyy}{College of Intelligence and Computing, Tianjin University, Tianjin, China}
\icmlaffiliation{lab}{Tianjin Key Lab of Machine Learning, Tianjin University, Tianjin, China}
\icmlaffiliation{comp1}{Institute of High Performance Computing (IHPC), Agency for Science, Technology and Research (A*STAR), Singapore}
\icmlaffiliation{comp2}{Centre for Frontier AI Research (CFAR), Agency for Science, Technology and Research (A*STAR), Singapore}
\icmlaffiliation{sch}{College of Computer Science, Sichuan University, Chengdu, China}

\icmlcorrespondingauthor{Changqing Zhang}{zhangchangqing@tju.edu.cn}

\icmlkeywords{Machine Learning, ICML}

\vskip 0.3in
]



\printAffiliationsAndNotice{}  

\begin{abstract} 
The inherent challenge of multimodal fusion is to precisely capture the cross-modal correlation and flexibly conduct cross-modal interaction. To fully release the value of each modality and mitigate the influence of low-quality multimodal data, dynamic multimodal fusion emerges as a promising learning paradigm. Despite its widespread use, theoretical justifications in this field are still notably lacking. \textit{Can we design a provably robust multimodal fusion method?}  This paper provides theoretical understandings to answer this question under a most popular multimodal fusion framework from the generalization perspective. We proceed to reveal that several uncertainty estimation solutions are naturally available to achieve robust multimodal fusion. Then a novel multimodal fusion framework termed Quality-aware Multimodal Fusion (QMF) is proposed, which can improve the performance in terms of classification accuracy and model robustness. Extensive experimental results on multiple benchmarks can support our findings.
\end{abstract}
\section{Introduction}
Our perception of the world is based on multiple modalities, e.g., touch, sight, hearing, smell and taste. With the development of sensory technology, we can easily collect diverse forms of data for analysis. For example, multi-sensor in autonomous driving and wearable electrical devices~\cite{xiao2020multimodal,wen2022wearable}, or various examinations in medical diagnosis and treatment~\cite{qiu2022multimodal,acosta2022multimodal}. Intuitively, fusing information from different modalities offers the possibility of exploring cross-modal correlation and gaining better performance. However, conventional fusion methods have largely overlooked the unreliable quality of multimodal data. In real-world settings, the quality of different modalities usually varies due to unexpected environmental issues. Some recent studies have shown both empirically and theoretically that multimodal fusion may fail on low-quality multimodal data, e.g., imbalanced~\cite{wang2020makes,peng2022balanced,huang2022modality}, noisy or even corrupted~\cite{huang2021learning} multimodal data. Empirically, it is recognized that multimodal models cannot always outperform unimodal models especially in a high noise ~\cite{scheunders2007wavelet,eitel2015multimodal,silva2022noise} or imbalanced modality quality~\cite{wu2022characterizing,peng2022balanced} regime. Theoretically, the previous study proves that the advantages of multimodal learning may vanish under the setting of limited data volume~\cite{huang2021makes} which implies the exploitation of cross-modal relationship is not a free lunch. To fully release the value of each modality and mitigate the influence of low-quality multimodal data, introducing dynamic fusion mechanism emerges as a promising way to obtain reliable predictions. As a concrete example, previous work~\cite{guan2019fusion} proposes a dynamic weighting mechanism to depict illumination condition of scenes. By introducing dynamics, they can integrate reliable cues from multi-spectral data for around-the-clock applications (e.g., pedestrian detection in security surveillance and autonomous driving). Dynamic fusion has been used in diverse real-world multimodal applications, including multimodal classification~\cite{han2021trusted, geng2021uncertainty, han2022multimodal}, regression~\cite{ma2021trustworthy}, object detection~\cite{li2022trustworthy, zhang2019weakly,chen2022multimodal} and semantic segmentation~\cite{tian2020uno}. While dynamic multimodal fusion shows excellent power in practice, theoretical understanding is notably lack in this field with the following fundamental open problem:
\textit{Can we realize reliable multimodal fusion in practice with theoretical guarantee?}

\begin{figure*}[!t]
    \centering
    \includegraphics[width=0.95\textwidth]{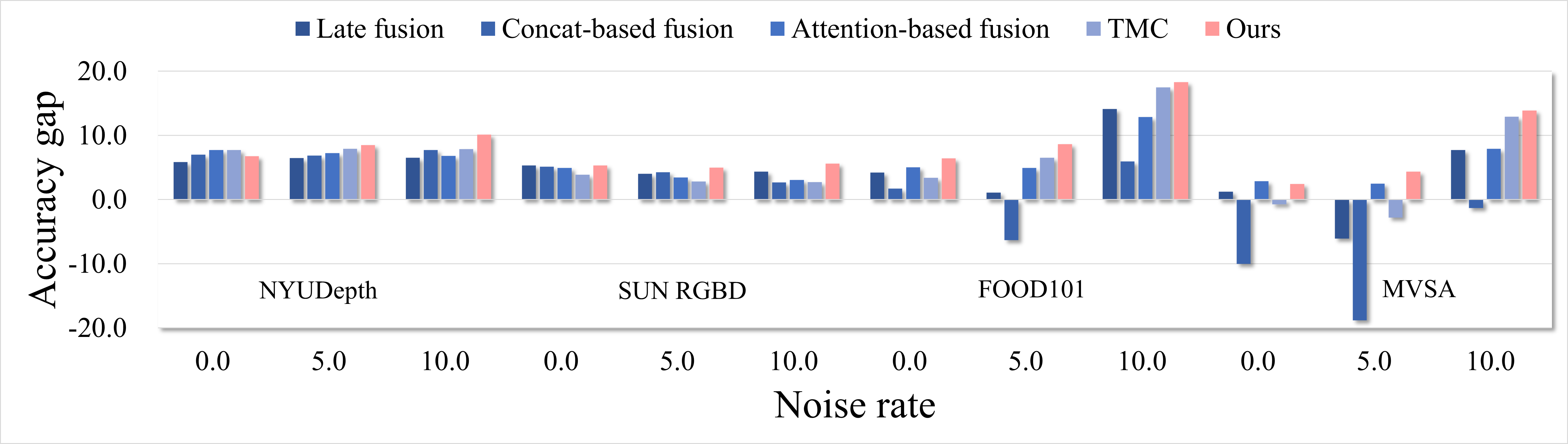}
    \caption{Visualization of accuracy gap between multimodal learning methods (e.g., late fusion, align-based fusion, MMTM) and single-modal learning methods using the best modality on noisy multimodal data. Noted that the performance existing multimodal fusion methods degrade significantly of compared with their best unimodal counterparts in a high noise regime, while the proposed QMF consistently outperforms unimodal methods on low-quality data.
} 
    \label{fig:robust}
\end{figure*}

\begin{figure*}[!t]
    \centering
    \includegraphics[width=0.99\textwidth]{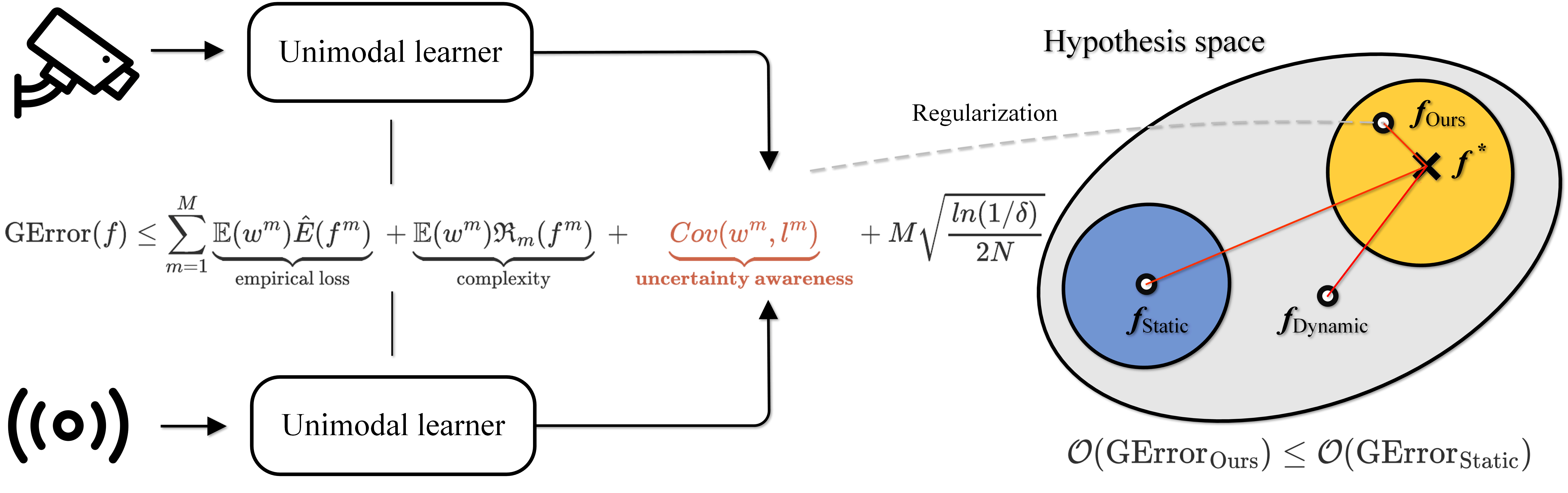}
    \caption{\textbf{Left:} The generalization error upper bound of multimodal fusion method $f$ can be characterized by its performance on each modality in terms of empirical loss, model complexity and uncertainty awareness. \textbf{Right:} Dynamic vs Static multimodal fusion hypothesis space, where the latter is a subset of the former. $f_{\text{static}}$, $f_{\text{dynamic}}$ are the hypothesises of static and dynamic fusion methods respectively and $f^*$ is the true mapping. Informally, closer to the true mapping leads to less error. Under some certain conditions, dynamic multimodal fusion methods (e.g., the proposed QMF) can be well regularized and thus provably achieve better generalization ability.} 
    \label{fig:framework}
\end{figure*}

This paper tries to shed light upon the theoretical advantage and criterion of robust multimodal fusion. Following previous works in multimodal learning theory~\cite{huang2021learning,wang2020makes}, the framework we study is also abstracted from decision-level multimodal fusion, which is one of the most fundamental research topics in multimodal learning~\cite{baltruvsaitis2018multimodal}. In particular, we devise a novel Quality-aware Multimodal Fusion (\textbf{QMF}) framework for multimodal learning. Key to our framework, we leverage energy-based uncertainty to characterize the quality of each modality. Our contributions can be summarised as follows:
\vspace{-0.3cm}
\begin{itemize}
\setlength{\itemsep}{1mm}
\setlength{\parskip}{0pt}
\item This paper provides a rigorous theoretical framework to understand the advantage and criterion of robust multimodal fusionas shown in Figure~\ref{fig:framework}. Firstly, we characterize the generalization error bound of decision-level multimodal fusion methods from a Rademacher complexity perspective. Then, we identify under what conditions dynamic fusion outperforms static, i.e., when the fusion weights of multimodal fusion is negatively correlates to the unimodal generalization errors, dynamic fusion methods provably outperform static.

\item Under the theoretical analysis, we proceed to reveal that the generalization ability of dynamic fusion coincides with the performance of uncertainty estimation. This directly implies a principle to design and evaluate new dynamic fusion algorithms.

\item Directly motivated by the above analysis, we propose a novel dynamic multimodal fusion method termed Quality-aware Multimodal Fusion (\textbf{QMF}), which serves as a realization for provably better generalization ability. As shown in Figure~\ref{fig:robust}, extensive experiments on commonly used benchmarks are carried out to empirically validate the theoretical observations.

\end{itemize}
\vspace{-0.3cm}

\section{Related works}
\subsection{Multimodal Fusion}
Multimodal fusion is one of the most original and fundamental topics in multimodal learning, which typically aims to integrate modality-wise features into a joint representation for downstream multimodal learning tasks. Multimodal fusion can be classified into early fusion, intermediate fusion and late fusion. Although studies in neuroscience and machine learning suggest that intermediate fusion could benefit representation learning~\cite{schroeder2005multisensory,macaluso2006multisensory}, late fusion is still the most widely used method for multimodal learning due to its interpretation and practical simplicity. By introducing modality-level dynamics based on various strategies, dynamic fusion practically improves overall performance. As a concrete example, the previous work~\cite{guan2019fusion} proposes a dynamic weighting mechanism to depict illumination conditions of scenes. By introducing dynamics, they can integrate reliable cues from multi-spectral data for around-the-clock applications (e.g., pedestrian detection in security surveillance and autonomous driving). Combining with additional dynamic mechanism (e.g., a simple weighting strategy or Dempster-Shafer Evidence Theory~\cite{shafer1976mathematical}), recent uncertainty-based multimodal fusion methods show remarkable advantages in various tasks, including clustering~\cite{geng2021uncertainty}, classification~\cite{han2021trusted,han2022multimodal,tellamekala2022cold,subedar2019uncertainty,chen2022uncertainty}, regression~\cite{ma2021trustworthy}, object detection~\cite{zhang2019weakly,li2022confidence} and semantic segmentation~\cite{tian2020uno,chang2022fast}. 

\subsection{Uncertainty Estimation}
\label{sec:uncertainty}
Multimodal machine learning has achieved great success in various real-world application. However, the reliability of current fusion methods is still notably unexplored, which limits their application in safety-critic field (e.g., financial risk, medical diagnosis). The motivation of uncertainty estimation is to indicate whether the predictions given by machine learning models are prone to be wrong. Many uncertainty estimation methods have been proposed in the past decades, including Bayesian neural networks (BNNs)~\cite{denker1990transforming, mackay1992bayesian, neal2012bayesian} and its varieties~\cite{gal2016dropout, han2022umix}, deep ensembles~\cite{lakshminarayanan2017simple, havasi2020training}, predictive confidence~\cite{hendrycks2016baseline}, Dempster-Shafer thoery~\cite{han2021trusted} and energy score~\cite{liu2020energy}. Predictive confidence expects the predicted class probability to be consistent with the empirical accuracy, which is usually referred in classification tasks. Dempster-Shafer theory (DST) is a generalization of Bayesian theory to subjective probabilities and a general framework for modeling epistemic uncertainty. Energy score emerges as a promising way to capture Out-of-Distribution (OOD) uncertainty, which arises when a machine learning model encounters an input that differs from its training data, and thus the output from the model is unreliable. A plethora of recent researches have studied the issue of OOD uncertainty~\cite{ming2022poem, chen2021atom, meinke2019towards, hendrycks2018deep}. In this paper, we investigate predictive confidence, the Dempster-Shafer theory and energy score due to their theoretical interpretability and effectiveness.
\section{Theory}
In this section, we first clarify the basic notations and the formal definition of multimodal fusion used in Section \ref{Sec:Preliminaries}. Then we provide main theoretical results in Section~\ref{Sec:Theory} to rigorously demonstrate when and how dynamic fusion methods work from the perspective of generalization ability~\cite{bartlett2002rademacher}. 
Due to space constraints, we defer the full details to Appendix~\ref{appendix-A} and only present a brief summary of the proofs.

\subsection{Preliminaries}\label{Sec:Preliminaries}
We initialize by introducing the necessary notations for our theoretical frameworks. Considering a learning task on the data $(x,y)\in\mathcal{X}\times \mathcal{Y}$, where $x=\{{x}^{(1)},\cdots,{x}^{(M)}\}$ has $M$ modalities and $y\in\mathcal{Y}$ denotes the data label. The multimodal training data is defined as $D_{\text{train}}=\{x_i,y_i\}_{i=1}^N$. Specifically, we use $\mathcal{X}$, $\mathcal{Y}$ and $\mathcal{Z}$ to denote the input space, target space and latent space. Similar to the previous work in multimodal learning theory~\cite{huang2021makes}, we define $h:\mathcal{X}\mapsto\mathcal{Z}$ is a multimodal fusion mapping from the input space to the latent space, and $g:\mathcal{Z}\mapsto\mathcal{Y}$ is a task mapping. Our goal is to learn a reliable multimodal model $f=g\circ h(x)$ performing well on the unknown multimodal test dataset $D_{\text{test}}$. $D_{\rm train}$ and $D_{\rm test}$ are both drawn from joint distribution $\mathcal{D}$ over $\mathcal{X}\times \mathcal{Y}$. Here $f=g\circ h(x)$ represents the composite function of $h$ and $g$.

\subsection{When and How Dynamic Multimodal Fusion Help}\label{Sec:Theory}
For simplicity, we provide analysis of ensemble-like late fusion strategy using logistic loss function in two-class classification setting. Our analysis follows this roadmap: (1) we first characterize the generalization error bound of dynamic late fusion using Rademacher complexity~\cite{bartlett2002rademacher} and then separate the bound into three components (Theorem~\ref{theorem:bound}); (2) base on above separation, we further prove that dynamic fusion achieves better generalization ability under certain conditions (Theorem~\ref{theorem:condition}). We initiate our analysis with the basic setting as follows.

\textbf{Basic setting.} Under a $M$ input modalities and two-class classification scenario, we define $f^m$ as the unimodal classifier on modality $x^{(m)}$. The final prediction of late fusion multimodal method is calculated by weighting decisions from different modalities: $ f(x)=\sum_{m=1}^M w^m\cdot f^m(x^{(m)})$, where $f(x)$ denotes the final prediction. In contrast to static late fusion, the weights in dynamic multimodal fusion are generated dynamically and vary for different samples. For clarity, we use subscript to distinguish them, i.e., $w^m_{\rm static}$ refers to the ensemble weight of modality $m$ in static late fusion and $w^m_{\rm dynamic}$ refers to the weight in dynamic fusion. Specifically, $w^m_{\rm static}$  is a constant and $w^m_{\rm dynamic}(\cdot)$ is a function of the input sample $x$. The generalization error of two-class multimodal classifier $f$ is defined as:
 \begin{equation}
\label{eq:gerror}
\text{GError}(f)=\mathbb{E}_{(x, y)\sim \mathcal{D}} [\ell(f(x),y)],
\end{equation}
where $\mathcal{D}$ is the unknown joint distribution, and $\ell$ is logistic loss function. For convenience, we simplify unimodal classifier loss $\ell(f^m(x^m),y)$ as $l^m$ and omit the inputs in the following analysis. Now we present our first main result regarding multi-modal fusion.

\begin{theorem}[Generalization Bound of Multimodal Fusion]\rm
\label{theorem:bound}
Let $D_{\text{train}}=\{x_i,y_i\}_{i=1}^N$ be a training dataset of $N$ samples, $\hat{E}(f^m)$ is the unimodal empirical errors of $f^m$ on $D_{\rm train}$. Then for any hypothesis $f$ in $\mathcal{H}$ (i.e., $\mathcal{H}:\mathcal{X}\rightarrow \{-1, 1\}$, $f\in \mathcal{H}$) and $1>\delta>0$, with probability at least $1-\delta$, it holds that
\begin{equation}
\begin{split}
\text{GError}(f)\leq\underbrace{\sum_{m=1}^M\mathbb{E}(w^m)\hat{E}(f^m)}_{\text{Term-L (average empirical loss)}}+\underbrace{\sum_{m=1}^M\mathbb{E}(w^m)\mathfrak{R}_m(f^m)}_{\text{Term-C (average complexity)}}\\+
\textcolor{mycolor3}{\underbrace{\sum^M_{m=1}Cov(w^m,l^m)}_{\text{Term-Cov (covariance)}}}+M\sqrt{\frac{ln(1/\delta)}{2N}},
\end{split}
\end{equation}
where $\mathbb{E}(w^m)$ is the expectations of fusion weights on joint distribution $\mathcal{D}$, $\mathfrak{R}_m(f^m)$ is Rademacher complexity, $Cov(w^m,\ell^m)$ is the covariance between fusion weight and loss.
\end{theorem}

Intuitively, Theorem~\ref{theorem:bound} demonstrates that the generalization error of multimodal classifier is bounded by the weighted average performances of all the unimodal classifiers in terms of empirical loss, model complexity and the covariance between fusion weight and unimodal loss. Having established the general error bound, our next goal is to verify when dynamic multimodal late fusion indeed achieves tighter bound than that of static late fusion. Informally, in Eq.~\ref{eq:gerror}, Term-Cov measures the joint variability of $w^m$ and $\ell^m$. Remember that in static multimodal fusion $w^m_{\rm static}$ is a constant, which means $\text{Term-Cov}=0$ for any static fusion methods. Thus the generalization error bound of static fusion methods reduces to
\begin{equation}
\begin{split}
\text{GError}(f_{\rm static})&\leq\underbrace{\sum_{m=1}^M w^m_{\text{static}}\hat{E}(f^m)}_{\text{Term-L (average empirical loss)}}\\&+\underbrace{\sum_{m=1}^Mw^m_{\text{static}} \mathfrak{R}_m(f^m)}_{\text{Term-C (average complexity)}}+M\sqrt{\frac{ln(1/\delta)}{2N}}.
\end{split}
\end{equation}
So when summation of $\text{Term-L}$ , $\text{Term-C}$ is invariant or smaller in dynamic fusion and $\text{Term-Cov}\leq 0$, we can ensure that dynamic fusion provably outperforms static fusion. This theorem is formally presented as

\begin{theorem}\rm
\label{theorem:condition}
Let $\mathcal{O}(\text{GError}(f_{\text{dynamic}}))$, $\mathcal{O}(\text{GError}(f_{\text{static}}))$ be the upper bound of generalization error of multimodal classifier using dynamic and static fusion strategy respectively. $\hat{E}(f^m)$ is the unimodal empirical errors of $f^m$ on $D_{\text{train}}$ defined in Theorem~\ref{theorem:bound}. Then for any hypothesis $f_{\text{dynamic}}$, $f_{\text{static}}$ in $\mathcal{H}:\mathcal{X}\rightarrow \{-1, 1\}$ and $1>\delta>0$, it holds that
\begin{equation}
    \mathcal{O}(\text{GError}(f_{\text{dynamic}}))\leq \mathcal{O}(\text{GError}(f_{\text{static}}))
\end{equation}
with probability at least $1-\delta$, if we have

\begin{equation}
\label{eq:expectation}
    \mathbb{E}(w^m_{\text{dynamic}})=w^m_{\text{static}}
\end{equation}

and
\begin{equation}
\label{eq:correlation}
    r(w^m_{\text{dynamic}},\ell(f^m))\leq 0
\end{equation}

for all input modalities, where $r$ is the Pearson correlation coefficient which measures the correlation between fusion weights $w^m_{\text{dynamic}}$ and unimodal loss $\ell^m$.
\end{theorem}

\textbf{Remark.} Theoretically, optimizing over the same function class efficiently results in the same empirical loss. Suppose for each modality $m$, the unimodal classifier $f^m$ we used in dynamic and static fusion are of the same architecture, then the intrinsic complexity of unimodal classifier $\mathfrak{R}_m(f^m)$ and empirical risk $\hat{E}(f^m)$ can be invariant. Thus in this case, it holds that

\begin{equation}
\label{eq:empirical}
    \sum_{m=1}^M\mathbb{E}(w^m_{\text{dynamic}})\hat{E}(f^m)\leq \sum_{m=1}^M w^m_{\text{static}}\hat{E}(f^m),
\end{equation}

and
\begin{equation}
\label{complexity}
    \sum_{m=1}^M\mathbb{E}(w^m_{\text{dynamic}})\mathfrak{R}_m(f^m)\leq\sum_{m=1}^M w^m_{\text{static}}\mathfrak{R}_m(f^m),
\end{equation}

if Eq.~\ref{eq:expectation} is satisfied for any modality $m$. According to Theorem~\ref{theorem:condition}, it is easy to derive the conclusion that the main challenge of achieving reliable dynamic multimodal fusion is to learn a reasonable $w^m_{\text{dynamic}}(x)$ for each modality that satisfies Eq.~\ref{eq:expectation} and Eq.~\ref{eq:correlation}.

\section{Method}\label{Sec:Realization}
    Now we proceed to answer "How to realize robust dynamic fusion?". In this section, we theoretically identify the connection between dynamic multimodal fusion and uncertainty estimation. Then, a unified dynamic multimodal fusion framework termed Quality-aware Multimodal Fusion (QMF) is proposed. We next show how to realize this framework in decision-level late fusion and classification tasks to support our findings.

\subsection{Coincidence with Uncertainty Estimation}

Firstly, we focus on how to satisfy Eq.~\ref{eq:correlation}. As we discuss in Section~\ref{sec:uncertainty}, the common motivation of various uncertainty estimation methods is to provide an indicator of whether the predictions given by models are prone to be wrong. This motivation is inherently close to obtaining weights that satisfy Eq.~\ref{eq:correlation}. We formulate this claim with the following assumption
\begin{framed}
\vspace{-0pt}
\begin{assumption}
 Given an effective uncertainty estimator $u^m:\mathcal{X}\rightarrow\mathbb{R}$ on modality $m$, the estimated uncertainty $u^m(x)$ is positively correlated with its modal-specific loss $\ell^m(x)$:
 $r(u^m, \ell^m(x))\geq 0$, 
 where $r$ is the Pearson correlation coefficient. 
 \label{assumption-main}
\end{assumption}
\vspace{-0pt}
\end{framed}

This insight offers opportunity to explore novel dynamic fusion methods provably outperform conventional static fusion methods. Similar to previous dynamic fusion methods~\cite{blundell2015weight,zhang2019weakly,han2022multimodal}, we deploy modal-level weighting strategy to introduce dynamics.

\textbf{Uncertainty-aware weighting.} The uncertainty-aware fusion weighting $w^m:\mathcal{X}\rightarrow \mathbb{R}$ is a function that linearly and negatively relates to the corresponding uncertainty
\begin{equation}
\label{eq:uaw}
    w^m(x)=\alpha^m\ u^m(x)+\beta^m,
\end{equation}
where $\alpha^m<0$, $\beta^m\geq0$ are modal-specific hyper-parameters. $u^m(x)$ is the uncertainty of modality $m$. By tuning hyper-parameters $\alpha^m$, $\beta^m$, we can ensure dynamic fusion weights satisfied Eq.~\ref{eq:expectation} and ~\ref{eq:correlation} simultaneously. This lemma is formally presented as \label{lemma_1}

\textbf{Lemma 1 (Satisfiability).} With Assumption~\ref{assumption-main}, for any $w^m_{\text{ static}}\in \mathbb{R}$, there always exist $\beta^m\in \mathbb{R}$ such that

\begin{equation}
    \label{lemma1}
    \mathbb{E}(w^m_{\text{dynamic}})=w^m_{\text{static}}, 
 r(w^m_{\text{dynamic}},\ell(f^m))\leq 0.
\end{equation}

Once we obtain the fusion weights, we can perform uncertainty-aware weighting fusion in decision-level according to the following rule
\begin{equation}
\label{eq:dynamic fusion}
    f(x)=\sum_{m=1}^M w^{m}(x)\cdot f^m(x),
\end{equation}
where $f^m(x)$ defined in Section~\ref{Sec:Theory} denotes unimodal prediction on modality $m$.

\begin{algorithm*}[ht]
\SetKwInOut{Input}{\textbf{Input}}
\SetKwInOut{Output}{\textbf{Output}}
 	\caption{Training Pseudo Code of Quality-aware Multimodal Fusion (QMF)}
 	 	\label{alg:QMF}
 		\Input{ Multimodal training dataset $D_{\text{train}}$, the number of sampling $\rm T$, hyperparameters $\lambda$, temperature parameters $\{\mathcal{T}^m\}_{m=1}^M$, unimodal predictors $\{f^m(\cdot)\}_{i=m}^M$;}
        
        \Output{ The multimodal classifier $f$;}
        
 		\For {each iteration}{
 		    Obtain training sample $(x_i, y_i)$ from dataset $D_{\text{train}}$ and the decisions on each modality $f^m(x)$;
       
            Calculate uncertainty-aware fusion weights $[w_i^1,\cdots,w_i^m]$ defined in Eq.~\ref{eq:uaw};
            
            Update the average training loss $\kappa_i^m$ of each modalities;

            Obtain the multimodal decision by weighting unimodal predictions dynamically according to Eq.~\ref{eq:dynamic fusion};
            
            Update model parameters of each unimodal predictor by minimizing $\mathcal{L_{\text{overall}}}$ in Eq.~\ref{eq:loss}.
           }

\end{algorithm*}

\subsection{Enhance Correlation by Additional Regularization}
With the above analysis, the core challenges of robust dynamic multimodal fusion present in Section.~\ref{Sec:Theory} have been reduced to obtain an effective uncertainty estimator in Assumption~\ref{assumption-main}. In our implementation, we leverage energy score~\cite{liu2020energy}, which is a widely accepted metric in the literature of uncertainty learning. Energy score~\footnote{While another line of previous works usually incorporate an auxiliary outlier dataset (e.g., random noised out-of-distribution data) during training for higher performance, for clarity and a strictly fair comparison, we conduct our experiments without the help of additional data.} bridges the gap between the Helmholtz free energy of a given data point and its density. For multimodal data, the density functions of different modalities can be estimated by the corresponding energy function:
\begin{equation}
log\ p(x^{(m)}) = -\text{Energy}(x;f^m)/\mathcal{T}^m -log \ Z^m,
\end{equation}
where $x^{(m)}$ is the $m$-th input modality and $f^m$ is the unimodal classification model. $\text{Energy} (\cdot)$ is the energy function and $Z^m$ is an intractable constant for all $x^m$. The above equation suggests that $-\text{Energy}(x^{(m)};f^m)$ is linearly aligned with density $p(x^{(m)})$. The energy score for the $m$-th modality of input $x$ can be calculated as
\begin{equation}
\text{Energy}(x^{(m)})=-\mathcal{T}^m\cdot log \sum^K_{k}e^{f_k^m(x^{(m)})/\mathcal{T}^m},
\end{equation}
where $f_k^m(x^{(m)})$ is the output logits of classifier $f^m$ corresponding to the $k$-th class label and $\mathcal{T}^m$ is a temperature parameter. Intuitively, more uniformly distributed prediction leads to higher estimated uncertainty.

However, it has been shown experimentally that the uncertainty estimated in this way without additional regularization is not well enough to satisfy our Assumption~\ref{assumption-main}. To address this, we propose a sampling-based regularization technology to enhance the original method in terms of correlation. The most simple and straightforward way to improve the correlation between estimated uncertainty and respective loss is to leverage the sample-wise loss during training stage as supervision information. However, due to the over-parameterization phenomenon of deep neural networks, the losses constantly reduce to zero during training. Inspired by recent works in Bayesian learning~\cite{maddox2019simple} and uncertainty estimation~\cite{moon2020confidence,han2022umix}, we propose to leverage the information from historical training trajectory to regularize the fusion weights. Specifically, given the $m$-th modality of a sample $(x_i,y_i)$, the training average loss for $x^m_i$ is calculated as:
\begin{equation}
    \kappa^m_i=\frac{1}{\rm T}\sum_{\rm t=T_s}^{\rm T_s+T}\ell(y_i,f^m_{\theta_t}(x_i)),
\end{equation}

where $f^m_{\theta_t}$ is the unimodal classifier on each iteration epoch $t$ with parameters $\theta_t$. After training $\rm T_s-1$ epochs, we sample $\rm T$ times and calculate the average training loss.

Empirically, recent works~\cite{geifman2018bias} shown that easy-to-classify samples are learned earlier during training compared to hard-to-classify samples (e.g., noise samples~\cite{arazo2019unsupervised}). It is desirable to regularize a dynamic fusion model by learning the following relationship during training
\begin{equation}
    \kappa^m_i\geq \kappa^m_j\iff w^m_i\leq w^m_j.
\end{equation}
We now present the full definition of our regularization term as follows
\begin{equation}
    \mathcal{L}_{\text{reg}}=max(0,g(w^m_i,w_j^m)(\kappa_i^m-\kappa_j^m)+|w_i^m-w_j^m|),
\end{equation}
where
\begin{equation}
    g(w_i^m,w_j^m)=\left\{
\begin{aligned}
1\ \text{if}\ w^m_i>w^m_j,\\
0\ \text{if}\ w^m_i=w^m_j,\\
-1\ \text{otherwise}.
\end{aligned}
\right.
\end{equation}
Inspired by multi-task learning, we define the total loss function as a summation of standard cross-entropy classification losses of multiple modalities and the regularization term
\begin{equation}
\label{eq:loss}
    \mathcal{L_{\text{overall}}}=\mathcal{L_{\text{CE}}}(y,f(x))+\sum_{m=1}^M\mathcal{L_{\text{CE}}}(y,f^m(x^m))+\lambda\mathcal{L}_{\text{reg}},
\end{equation}
where $\lambda$ is a hyperparamter which controls the strength of regularization, $\mathcal{L}_{\rm CE}$ and $\mathcal{L}_{\rm reg}$ are the cross-entropy loss and reguralization term respectively. The whole training process is shown in Algorithm ~\ref{alg:QMF}.

\textbf{Intuitive explanation of the effectiveness of QMF.} Without loss of generality, we assume modality $x^A$ is clean and modality $x^B$ is noisy due to unknown environmental factors or sensor failure. At this time, $x^A$ is in the distribution of clean training data but $x^B$ deviates significantly from it. Accordingly, we have $u(x^A)\leq u(x^B)$ and thus $w^A\geq w^B$. Therefore, for our QMF, the multimodal decision will tend to more rely on the high-quality modality $x^A$ than the other modality $x^B$. By dynamically determining the fusion weights of each modality, the influence of the unreliable modalities can be alleviated.

\section{Experiment}
In this section, we conduct experiments on multiple datasets of diverse applications~\footnote{Code is available at \href{https://github.com/QingyangZhang/QMF}{https://github.com/QingyangZhang/QMF}.}. The main questions to be verified are highlighted here: 
\begin{itemize}
    \item Q1 Effectiveness I. Does the proposed method has better generalization ability than its counterparts? (Support Theorem~\ref{theorem:bound}) \item Q2 Effectiveness II. Under what conditions does uncertainty-aware dynamic multimodal fusion work? (Support Theorem~\ref{theorem:condition}) \item Q3 Reliability. Does the proposed method have an effective perception for the uncertainty of modality? (Support Assumption~\ref{assumption-main}) \item Q4 Ablation study. What is the key factor of performance improvement in our method?
\end{itemize}

\subsection{Experimental Setup}
We briefly present the experimental setup here, including the experimental datasets and comparison methods.  Please refer to Appendix~\ref{appendix-B} for more detailed setup.

\begin{table*}[!h]

\setlength{\abovecaptionskip}{0pt}
\begin{spacing}{1.35}
\caption{Classification comparison when 50\% of the modalities are corrupted with Gaussian noise i.e., zero mean with variance of $\epsilon$. The best three results are in bold brown and the best results are highlighted in bold blue. Full results with standard deviation are in Appendix. \label{tab:classification}}
\end{spacing}
\center
{
\renewcommand{\arraystretch}{1.25}
\begin{tabular}{c|cccccccc}
\toprule
\multirow{2}{*}{\text{Dataset}} & \multirow{2}{*}{\text{Dynamic}} & \multirow{2}{*}{\text{Method}} & \multicolumn{2}{c}{\text{$\mathbf{\epsilon=0.0}$}}& \multicolumn{2}{c}{\text{$\mathbf{\epsilon=5.0}$}}& \multicolumn{2}{c}{\text{$\mathbf{\epsilon=10.0}$}}
\\
 & & &Avg. &Worst. &Avg. &Worst. &Avg. &Worst.\\

\midrule
 \multirow{8}{*} {\text{\shortstack{NYU \\Depth V2}}} 
&\xmark     &  \text{RGB}                                        &${63.30}$ &${62.54}$  &${53.12}$ &${50.31}$
       &${45.46}$   &${42.20}$
       
        \\
&\xmark     &  \text{Depth} &${62.65}$ &${61.01}$  &${50.95}$ &${42.81}$ &${44.13}$ &${35.93}$ 
    
    \\ &\xmark  & {\text{Late fusion}}  &${69.14}$ &${68.35}$  &${59.63}$ &${53.98}$ &${51.99}$ &${44.95}$

    \\&\xmark   & {\text{Concat}} &${70.30}$ &$\textcolor{mycolor1}{\pmb{69.42}}$  &${59.97}$ &${55.89}$ &$\textcolor{mycolor1}{\pmb{53.20}}$ &$\textcolor{mycolor1}{\pmb{47.71}}$

    \\ &\xmark  & {\text{Align}} &$\textcolor{mycolor1}{\pmb{70.31}}$ &${68.50}$  &${59.47}$ &${56.27}$ &${51.74}$ &${44.19}$ 
       
    \\&\cmark   & {\text{MMTM}} &$\textcolor{mycolor1}{\pmb{71.04}}$ &$\textcolor{mycolor4}{\pmb{70.18}}$  &$\textcolor{mycolor1}{\pmb{60.37}}$ &$\textcolor{mycolor1}{\pmb{56.73}}$ &${52.28}$ &${46.18}$                                             
    \\&\cmark    & {\text{TMC}} &$\textcolor{mycolor4}{\pmb{71.06}}$ &$\textcolor{mycolor1}{\pmb{69.57}}$  &$\textcolor{mycolor1}{\pmb{61.04}}$ &${\textcolor{mycolor1}{\pmb{58.72}}}$ &$\textcolor{mycolor1}{\pmb{53.36}}$ &$\textcolor{mycolor1}{\pmb{49.23}}$                                            
    \\ \cline{2-9} &\cmark  & {\text{Ours}} &${70.09}$ &${68.81}$  &$\textcolor{mycolor4}{\pmb{61.62}}$ &$\textcolor{mycolor4}{\pmb{58.87}}$ &$\textcolor{mycolor4}{\pmb{55.60}}$ &$\textcolor{mycolor4}{\pmb{51.07}}$

        \\ \midrule  \multirow{8}{*} {\text{\shortstack{SUN\\RGB-D}}} 
     &\xmark & \text{RGB} &${56.78}$ &${56.51}$  &${48.40}$ &${47.16}$ &${42.94}$ &${41.02}$
    
    \\ &\xmark & \text{Depth}    &${52.99}$ &${51.32}$  &${37.81}$ &${35.63}$ &${33.07}$ &${30.41}$     

    \\   &\xmark & {\text{Late fusion}}     &$\textcolor{mycolor1}{\pmb{62.09}}$ &${60.55}$  &$\textcolor{mycolor1}{\pmb{52.44}}$ &$\textcolor{mycolor1}{\pmb{50.83}}$ &$\textcolor{mycolor1}{\pmb{47.33}}$ &$\textcolor{mycolor1}{\pmb{44.60}}$
    
    \\   &\xmark & {\text{Concat}} &$\textcolor{mycolor1}{\pmb{61.90}}$ &$\textcolor{mycolor1}{\pmb{61.19}}$  &$\textcolor{mycolor1}{\pmb{52.69}}$ &${50.61}$ &${45.64}$ &${42.95}$
    
    \\   &\xmark & {\text{Align}}  &${61.12}$ &${60.12}$  &${50.05}$ &${47.63}$ &${44.19}$ &${38.12}$                                
    
    \\   &\cmark & {\text{MMTM}}  &${61.72}$ &$\textcolor{mycolor1}{\pmb{60.94}}$  &${51.86}$ &$\textcolor{mycolor1}{\pmb{50.80}}$ &$\textcolor{mycolor1}{\pmb{46.03}}$ &$\textcolor{mycolor1}{\pmb{44.28}}$                                 
    
    \\   &\cmark & {\text{TMC}}  &${60.68}$ &${60.31}$  &${51.24}$ &${49.45}$ &${45.66}$ &${41.60}$                                     
    \\  \cline{2-9}  &\cmark & {\text{Ours}} &$\textcolor{mycolor4}{\pmb{62.09}}$ &$\textcolor{mycolor4}{\pmb{61.30}}$  &$\textcolor{mycolor4}{\pmb{53.40}}$ &$\textcolor{mycolor4}{\pmb{52.07}}$ &$\textcolor{mycolor4}{\pmb{48.58}}$ &$\textcolor{mycolor4}{\pmb{47.50}}$
    
\\ \midrule  \multirow{9}{*} {\text{\shortstack{FOOD\\101}}} 
 &\xmark &\text{Bow}  &${82.50}$  &${82.32}$  &${61.68}$ &${60.98}$   &${41.95}$ &${41.41}$
\\ &\xmark & \text{Img}   &${64.62}$ &${64.22}$   &${34.72}$ &${34.19}$  &${33.03}$ &${32.67}$   
\\ &\xmark & \text{Bert}   &${86.46}$ &${86.42}$   &${67.38}$  &${67.19}$ &${43.88}$  &${43.56}$ 
        
\\  &\xmark & {\text{Late fusion}} &$\textcolor{mycolor1}{\pmb{90.69}}$ &$\textcolor{mycolor1}{\pmb{90.58}}$  &${68.49}$  &${65.05}$ &$\textcolor{mycolor1}{\pmb{58.00}}$   &${55.77}$   
         
        \\&\xmark   & {\text{ConcatBow}}                                             &${70.77}$ &${70.68}$ &${38.28}$ &${37.95}$ &${35.68}$  &${34.92}$   
        
        \\&\xmark   & {\text{ConcatBert}} &${88.20}$ &${87.81}$ &${61.10}$  &${59.25}$ &${49.86}$    &${47.79}$                                           
       
        \\&\cmark   & {\text{MMBT}}&$\textcolor{mycolor1}{\pmb{91.52}}$ &$\textcolor{mycolor1}{\pmb{91.38}}$ &$\textcolor{mycolor1}{\pmb{72.32}}$  &$\textcolor{mycolor1}{\pmb{71.78}}$ &${56.75}$    &$\textcolor{mycolor1}{\pmb{56.21}}$ 
        
    \\&\cmark    & {\text{TMC}} &${89.86}$ &${89.80}$ &$\textcolor{mycolor1}{\pmb{73.93}}$  &$\textcolor{mycolor1}{\pmb{73.64}}$ &$\textcolor{mycolor1}{\pmb{61.37}}$ &$\textcolor{mycolor1}{\pmb{61.10}}$ 
        
       \\ \cline{2-9}  &\cmark   & {\text{Ours}}                                             &$\textcolor{mycolor4}{\pmb{92.92}}$ &$\textcolor{mycolor4}{\pmb{92.72}}$ &$\textcolor{mycolor4}{\pmb{76.03}}$ &$\textcolor{mycolor4}{\pmb{74.68}}$ &$\textcolor{mycolor4}{\pmb{62.21}}$   &$\textcolor{mycolor4}{\pmb{61.76}}$   
        \\ \midrule  \multirow{9}{*} {\text{MVSA}} 
        
&\xmark &  \text{Bow}  &${48.79}$   &${35.45}$  &${42.20}$   &${32.56}$ &${41.57}$   &${32.18}$ \\ 
&\xmark &  \text{Img}   &${64.12}$ &${62.04}$ &${49.36}$ &${45.67}$ &${45.00}$ &${39.31}$ \\
&\xmark &  \text{Bert}  &${75.61}$   &$\textcolor{mycolor1}{\pmb{74.76}}$  &$\textcolor{mycolor1}{\pmb{69.50}}$   &$\textcolor{mycolor1}{\pmb{65.70}}$ &${47.41}$   &${45.86}$\\ 
&\xmark &  \text{Late fusion}  &$\textcolor{mycolor1}{\pmb{76.88}}$   &${74.76}$  &${63.46}$   &${58.57}$ &${55.16}$   &${47.78}$\\ 
&\xmark &  \text{ConcatBow}  &${64.09}$   &${62.04}$  &${49.95}$   &${45.28}$ &${45.40}$   &${40.95}$\\ 
&\xmark &  \text{ConcatBert}  &${65.59}$   &${64.74}$  &${50.70}$   &${44.70}$ &${46.12}$   &${41.81}$\\ 
&\cmark &  \text{MMBT}  &$\textcolor{mycolor4}{\pmb{78.50}}$   &$\textcolor{mycolor4}{\pmb{78.04}}$  &$\textcolor{mycolor1}{\pmb{71.99}}$   &$\textcolor{mycolor1}{\pmb{69.94}}$ &$\textcolor{mycolor1}{\pmb{55.35}}$   &$\textcolor{mycolor1}{\pmb{52.22}}$\\ 
&\cmark &  \text{TMC}  &${74.88}$   &${71.10}$  &${66.72}$   &$\textcolor{mycolor1}{\pmb{60.12}}$ &${60.36}$   &$\textcolor{mycolor1}{\pmb{53.37}}$\\ 
 \cline{2-9} &\cmark  & {\text{Ours}} &$\textcolor{mycolor1}{\pmb{78.07}}$   &$\textcolor{mycolor1}{\pmb{76.30}}$  &$\textcolor{mycolor4}{\pmb{73.85}}$   &$\textcolor{mycolor4}{\pmb{71.10}}$ &$\textcolor{mycolor4}{\pmb{61.28}}$   &$\textcolor{mycolor4}{\pmb{57.61}}$\\

\bottomrule
\end{tabular}}
\vspace{0mm}
\end{table*}

\begin{figure*}[htbp]
  \centering
  \subfigure[NYU Depth V2]{
  \includegraphics[width=0.48\textwidth]{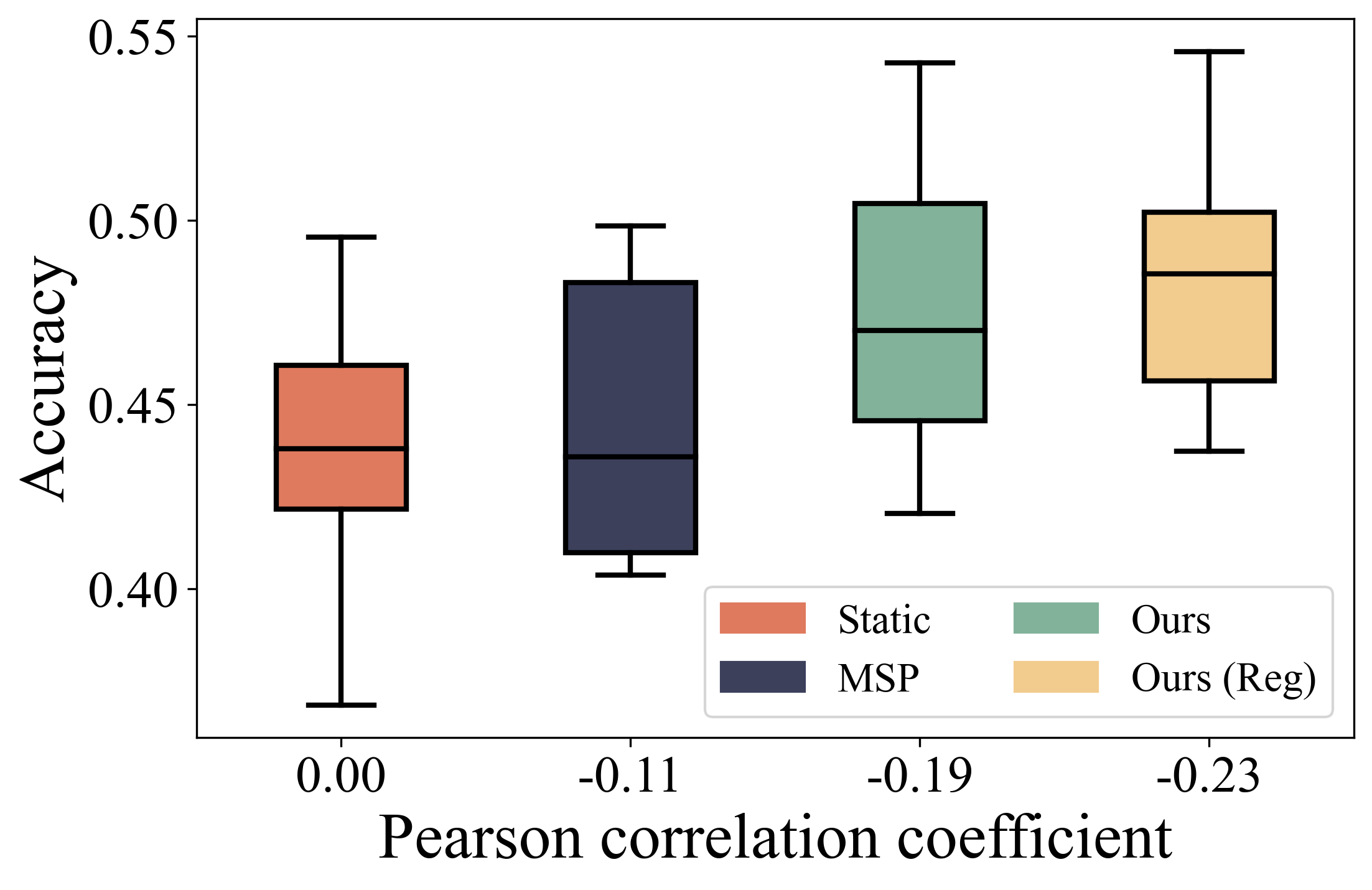}
  }
    \subfigure[SUN RGB-D]{
  \includegraphics[width=0.48\textwidth]{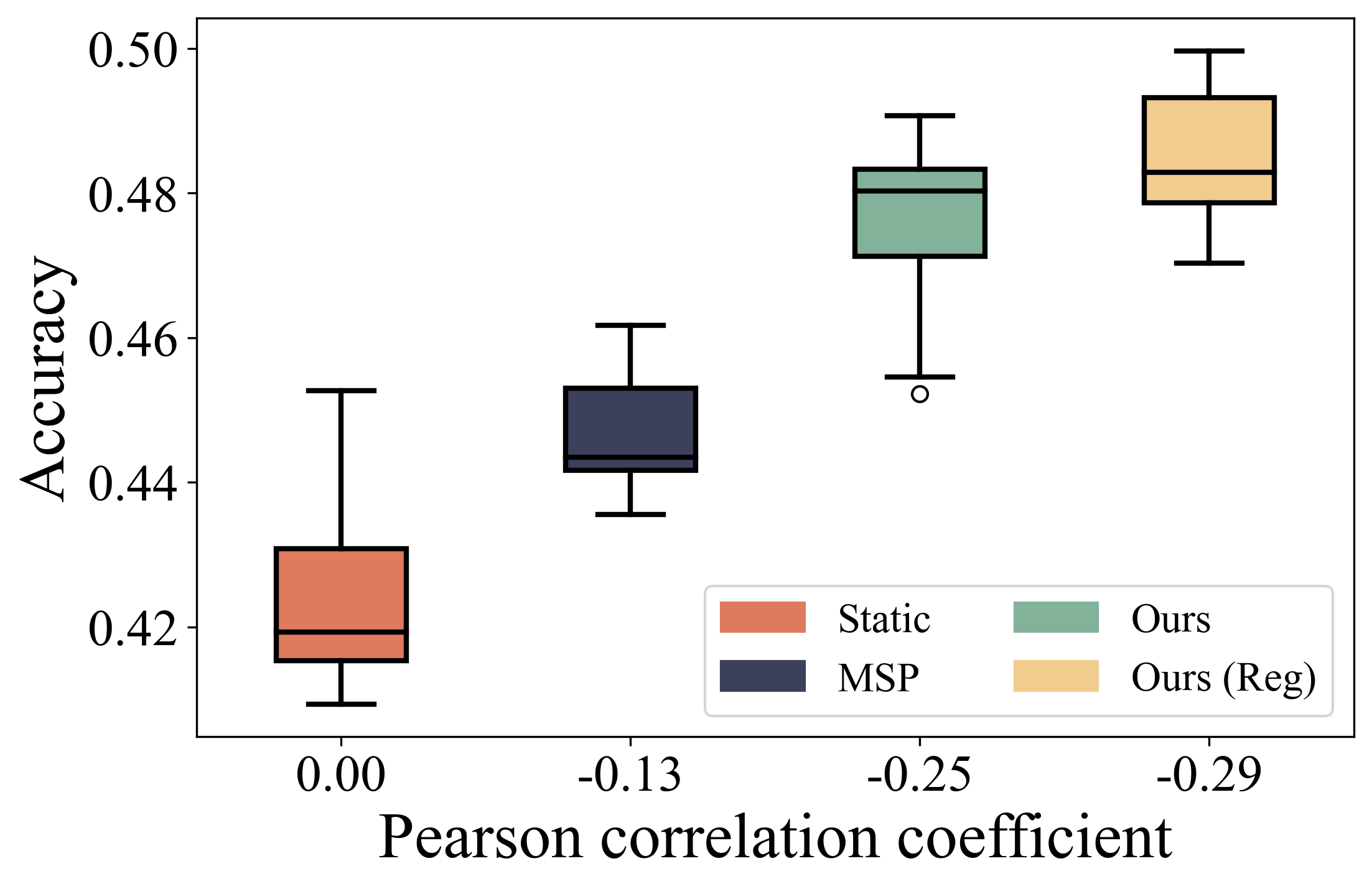}
  }
 \caption{Test accuracy and Pearson correlation coefficient achieved by different fusion methods over 10 times random experiments. The average and worst-case accuracy are highly consistency with uncertainty estimation ability.}
  \label{fig:exp2}
  \vspace{-0.0cm}
\end{figure*}

\begin{table*}[!htbp]
\vspace{-0.0cm}
\begin{center}
\begin{spacing}{1.35}   
\caption{Ablation study on NYU Depth V2. Full results with standard deviation are in Appendix C.1. \label{tab:ablation}}
\end{spacing}
\center
{
\setlength{\tabcolsep}{1.9mm}
\begin{tabular}{cccccccccc}
\toprule
\multirow{2}{*}{\text{UAW}}  & \multirow{2}{*}{\text{$\mathcal{L}_{\text{reg}}$}}  &
\multicolumn{2}{c}{$\mathbf{\epsilon=0.0}$} & \multicolumn{2}{c}{$\mathbf{\epsilon=5.0}$}& \multicolumn{2}{c}{$\mathbf{\epsilon=10.0}$}& \multicolumn{2}{c}{$\mathbf{\epsilon=20.0}$}
\\
 & &Avg. &Worst.  &Avg. &Worst.  &Avg. &Worst.  &Avg. &Worst.\\
\midrule
  
\xmark         &\xmark                                          &${69.14}$    &$68.35$
       &${59.62}$    &$53.98$
       &${51.94}$    &$44.95$
       &${43.76}$    &$36.85$ \\
                                                                             \xmark  &\cmark             
                                                         &${69.68}$  &${67.74}$
       &${61.35}$   &${58.26}$
       &${55.44}$   &$\textcolor{mycolor4}{\pmb{51.53}}$
       &${47.32}$    &${42.97}$                                \\  \cmark  &\xmark             
                                                         &${70.06}$  &$\textcolor{mycolor4}{\pmb{69.11}}$ 
                                                         &${61.59}$   &$57.49$
                                                         &${55.14}$ &$50.15$
                                                         &${47.46}$  &${42.05}$ \\\midrule \cmark  & \cmark            
                                                         &$\textcolor{mycolor4}{\pmb{70.09}}$   &${68.81}$
                                                         &$\textcolor{mycolor4}{{\pmb{61.62}}}$  &$\textcolor{mycolor4}{\pmb{58.87}}$
                                                         &$\textcolor{mycolor4}{{\pmb{55.81}}}$ &${51.07}$
                                                         &$\textcolor{mycolor4}{{\pmb{48.26}}}$ &$\textcolor{mycolor4}{\pmb{43.73}}$                                  \\ \bottomrule
\end{tabular}}
\end{center}
\vspace{-0.0cm}

\end{table*}

\begin{table}[t]
\begin{center}
\begin{spacing}{1.35}   
\caption{Pearson correlation coefficient $r$ between losses and fusion weights of test samples (a higher $|r|$ indicates a better uncertainty estimation). \label{tab:rel}}
\end{spacing}
\center
{
\setlength{\tabcolsep}{1.9mm}
\begin{tabular}{cccc}
\toprule
\multicolumn{1}{c}{}  &
\multicolumn{1}{c}{$\mathbf{\epsilon=0.0}$} & \multicolumn{1}{c}{$\mathbf{\epsilon=5.0}$}& \multicolumn{1}{c}{$\mathbf{\epsilon=10.0}$}
\\
\midrule

MSP      &${0.391}$    &${0.433}$  &${0.486}$       \\
Energy score     &${0.272}$  &${0.429}$   &${0.510}$  \\
Entropy       &${0.397}$  &${0.420}$ &${0.452}$       \\
Evidence      &${0.157}$ &${0.136}$   &${0.265}$              
\\\midrule Ours     &$\textcolor{mycolor4}{\pmb{{0.498}}}$ &$\textcolor{mycolor4}{\pmb{{{0.652}}}}$ &$\textcolor{mycolor4}{\pmb{{{0.735}}}}$              
\\ \bottomrule
\end{tabular}}
\end{center}
\vspace{-0.1cm}
\end{table}

\textbf{Tasks and datasets.} We evaluate our method on two multimodal classification tasks. $\circ$ Scenes Recognition: NYU Depth V2~\cite{silberman2012indoor} and SUN RGB-D~\cite{song2015sun} are two public indoor scenes recognition datasets, which are associated with two modalities, i.e., RGB and depth images. $\circ$ Image-text classification: The UPMC FOOD101 dataset~\cite{wang2015recipe} contains (possibly noisy) images obtained by Google Image Search and corresponding textual descriptions. MVSA sentiment analysis dataset~\cite{niu2016sentiment} includes a set of image-text pairs with manual annotations collected from social media. Although the datasets above are all under
the condition that $M=2$, it is intuitive and easy to generalize to $M\geq3$.

\textbf{Evaluation metrics.} Due to the randomness involved, we report the mean accuracy, standard deviation and worst-case accuracy on NYU Depth V2 and SUN RGB-D over 10 different seeds. To be consistent with existing works~\cite{han2022trusted, kiela2019supervised, yadav2023deep}, we repeat experiments over 3 times on UMPC FOOD101 and 5 times on MVSA.

\textbf{Compared methods.} For scene recognition task, we compare the proposed method with three static fusion methods: Late fusion, Concatenate-based fusion, Alignment-based fusion methods~\cite{wang2016learning} and two representative dynamic fusion methods, i.e., MMTM~\cite{joze2020mmtm} and TMC\footnote{There are two variants in~\cite{han2021trusted}: TMC and ETMC (with additional concatenated-based multimodal fusion strategy). TMC has comparable performance and is a more fair comparison.}~\cite{han2021trusted}. For image-text classification, we compare against strong unimodal baselines (i.e., Bow, Bert and ResNet-152) as well as sophisticated multimodal fusion methods, including Late fusion, ConcatBow, ConcatBERT and recent sota MMBT~\cite{kiela2019supervised}.

\subsection{Experimental Results}

\textbf{Classification robustness (Q1).}
To validate the robustness of the uncertainty-aware weighting fusion, we evaluate QMF and the compared methods in terms of average and worst-case accuracy under Gaussian noise (for image modality) and blank noise (for text modality) following previous works~\cite{han2021trusted,ma2021trustworthy,verma2021graphmix,hu2019learning,xie2017data}. More results under different types of noise (e.g. Salt-Pepper Noise) can be found in Appendix~\ref{sec:app-C.2}. The experimental results are presented in Table~\ref{tab:classification}. It is observed that QMF usually performs in the top three in terms of both average and worst-case accuracy. This observation indicates that QMF has better generalization ability than their counterparts experimentally. It is also noteworthy that the QMF outperforms the prior \textbf{state-of-the-art} methods (i.e., MMBT and TMC) on large-scale benchmark UPMC FOOD101, which illustrates the superiority of the proposed method.

\textbf{Connection to uncertainty estimation (Q2).}
We further conduct comparisons with QMF realized by various uncertainty estimation algorithms, i.e., prediction confidence~\cite{hendrycks2016baseline} and Dempster-Shafer evidence theory (DST)~\cite{han2021trusted}. According to comparison results shown in the Figure~\ref{fig:exp2}, it is clear that (i) the generalization ability (i.e., average and worst-case accuracy) of dynamic fusion methods coincide with their uncertainty estimation ability and (ii) our QMF achieves the best performance in terms of classification accuracy and uncertainty estimation in the meantime. This comparison reveals the underlying reason of why QMF outperforms other fusion methods and supports Theorem~\ref{theorem:condition}. We show the results on NYU Depth V2 and SUN RGB-D under Gaussian noise with zero mean and variance of 10.

\textbf{Reliability of QMF (Q3).} We calculate the fusion weights defined in Eq.~\ref{eq:uaw} of different modalities in Table~\ref{tab:rel} on UPMC FOOD-101. It is observed that the fusion weights of QMF have the most effective perception of modal quality compared with other uncertainty estimation methods (in terms of correlation). This observation justifies our expectation of uncertainty-aware weights in Eq.~\ref{eq:uaw}.

\textbf{Ablation study (Q4).}
We compare different combinations of components (i.e., uncertainty-aware weighting and the regularization term $\mathcal{L}_{\text{reg}}$). Here we also employ Gaussian noise on NYU Depth V2 in Table~\ref{tab:ablation}, and more results can be found in the Appendix C.1. It is easy to conclude that 1) adding $\rm \mathcal{L}_{reg}$ is beneficial to obtaining more reasonable fusion weights; 2) the best performance could be expected with the full QMF. Please refer to Table.~\ref{tab:appendix-ablation} in Appendix C.1 for full results with standard deviation.

In summary, the empirical results can support our theoretical findings. These works identify the causes and conditions of performance gains of dynamic multimodal fusion methods. The proposed method can help to improve robustness on multiple datasets.

\section{Limitations}

Even though the proposed method achieves superior performance, there are still some potential limitations. Firstly, the fusion weights of QMF are based on uncertainty estimation, which can be a challenging task in the real world. For example, in our experiments, we can only achieve mild Pearson's $r$ on NYU Depth V2 and SUN RGB-D dataset. Therefore, it is important and valuable to explore novel uncertainty estimation methods in the future work. Secondly, though we characterize the generalization error bound of the proposed method, our theoretical justifications are based on Assumption~\ref{assumption-main}. However, previous work~\cite{fang2022out} reveals that OOD detection is not learnable under some scenarios. Thus it’s still a challenging open problem to further characterize the generalization ability of dynamic multimodal fusion.

\section{Conclusions and Future works}
Introducing dynamics in multimodal fusion has yielded remarkable empirical results in various applications, including image classification, object detection and semantic segmentation. Many state-of-the-art multimodal models introduce dynamic fusion strategies, but the inductive bias provided by this technique is not well understood. In this paper, we provide rigorous analysis towards understanding when and what dynamic multimodal fusion methods are more robust on multimodal data in the wild. These findings demonstrate the connection between uncertainty learning and robust multimodal fusion, which further implies a principle to design novel dynamic multimodal fusion methods. Finally, we perform extensive experiments on multiple benchmarks to support our findings. In the work, the energy-based weighting strategy is devised, and other uncertainty estimation ways could be explored. Another interesting direction is proving the dynamic fusion under a more general setting.

\section*{Acknowledgments}
This work is partially supported by the National Natural Science Foundation of China (Grant No. 61976151) and A*STAR Central Research Fund. We gratefully acknowledge the support of MindSpore and CAAI. The authors would like to thank Zhipeng Liang (Hong Kong University of Science and Technology) for checking on math details and Zongbo Han, Huan Ma (Tianjin University) for their comments on writing. The authors also appreciate the suggestions from ICML anonymous peer reviewers.

\bibliography{example_paper}
\bibliographystyle{icml2023}

\appendix

\newpage
\appendix
\onecolumn
\addcontentsline{toc}{section}{Appendix} 
\section*{Appendix}

\section{Proofs}
\label{appendix-A}
\subsection{Proof of Theorem~\ref{theorem:bound}}
\begin{proof}
Let $(x,y)\sim \mathcal{D}$ denotes the multimodal sample, then we have
\begin{equation}
    \ell(f(x),y)=\ell(\sum_{m=1}^M w^mf^m(x^{(m)}),y).
\end{equation}

Noted that $\ell$ is a convex logistic loss function, which indicates that
\begin{equation}
    \ell(f(x),y)=\ell(\sum_{m=1}^M w^mf^m(x^{(m)}),y)\leq \sum_{m=1}^M w^m\ell(f^m(x^{(m)}),y).
\end{equation}
Then we take the expectation on both sides of the above equation
\begin{equation}
    \mathbb{E}_{(x,y)\sim \mathcal{D}}\ell(f(x),y)\leq \mathbb{E}_{(x,y)\sim \mathcal{D}}\sum_{m=1}^M w^m\ell(f^m(x^{(m)}),y),
\end{equation}
since expectation is a linear operator and the expected value of the product is equal to the product of the expected values plus the covariance, we can further decompose the right-hand side of the equation into

\begin{align} 
\mathbb{E}_{(x,y)\sim \mathcal{D}}\ell(f,y)&\leq \sum_{m=1}^M\mathbb{E}_{(x,y)\sim \mathcal{D}}[ w^m\ell(f^m,y)]\\ &=\sum_{m=1}^M \mathbb{E}_{(x,y)\sim \mathcal{D}}(w^m)\mathbb{E}_{(x,y)\sim \mathcal{D}}(\ell(f^m,y))+Cov(w^m,\ell(f^m,y))
\end{align}
Next, we recap the Rademacher complexity measure for model complexity. We use complexity-based learning theory~\cite{bartlett2002rademacher} (Theorem 8) to quantify the generalization error of unimodal models.

Let $D_{\rm train}=\{x_i,y_i\}_{i=1}^N$ be the training dataset of $N$ samples, $\hat{E}(f^m)$ is the unimodal empirical error of $f^m$. Then for any hypothesis $f^m$ in $\mathcal{H}$ (i.e., $\mathcal{H}:\mathcal{X}\rightarrow \{-1, 1\}$, $f\in \mathcal{H}$) and $1>\delta>0$, with probability at least $1-\delta$, we have
\begin{align}
\nonumber
\mathbb{E}_{(x,y)\sim \mathcal{D}}(f^m)\leq \hat{E}(f^m)+\mathfrak{R}_m(\mathcal{H})+\sqrt{\frac{ln(1/\delta)}{2N}},
\end{align}
where $\mathfrak{R}_m(f^m)$ is the Rademacher complexities. 

Finally, it holds with probability at least $1-\delta$ that
\begin{equation}
    \text{GError}(f)\leq \sum_{m=1}^M\mathbb{E}(w^m)\hat{E}(f^m)+\mathbb{E}(w^m)\mathfrak{R}_m(\mathcal{H})+Cov(w^m,\ell(f^m,y))+M\sqrt{\frac{ln(1/\delta)}{2N}}.
\end{equation}
\end{proof}

\subsection{Proof of Theorem~\ref{theorem:condition}}

\begin{proof}
Let $\mathcal{O}(\text{GError}(f_{\text{dynamic}}))$, $\mathcal{O}(\text{GError}(f_{\text{static}}))$ be the upper bound of generalization error of multimodal classifier using dynamic and static fusion strategy respectively, $\hat{E}(f^m)$ is the unimodal empirical errors of $f^m$ on $D_{\text{train}}$ defined in Theorem.~\ref{theorem:bound}. Theoretically, optimizing over the
same function class results in the same empirical risk. Therefore
\begin{equation}
    \hat{E}(f_{\rm static}^m)=\hat{E}(f_{\rm dynamic}^m).
\end{equation}
Additionally, the intrinsic complexity of unimodal classifier $\mathfrak{R}_m(f^m)$ is also invariant
\begin{equation}
    \mathfrak{R}_m(f^m_{\rm static})=\mathfrak{R}_m(f^m_{\rm dynamic}).
\end{equation}
Thus in this special case, it holds that

\begin{equation}
\label{appendix-eq:empirical}
    \sum_{m=1}^M\mathbb{E}(w^m_{\text{dynamic}})\hat{E}(f^m)\leq \sum_{m=1}^M w^m_{\text{static}}\hat{E}(f^m),
\end{equation}
and
\begin{equation}
\label{appdendix-complexity}
    \sum_{m=1}^M\mathbb{E}(w^m_{\text{dynamic}})\mathfrak{R}_m(f^m)\leq\sum_{m=1}^M w^m_{\text{static}}\mathfrak{R}_m(f^m),
\end{equation}

if $\mathbb{E}(w^m_{\text{dynamic}})=w^m_{\text{static}}$.

Since the covariance and correlation coefficient have the same sign, when $r(w^m,l^m)\leq 0$, the covariance $Cov(w^m,l^m)$ is also less than or equal to 0. Therefore, it holds that
\begin{equation}
    \mathcal{O}(\text{GError}(f_{\text{dynamic}}))\leq \mathcal{O}(\text{GError}(f_{\text{static}}))
\end{equation}
with probability at least $1-\delta$, if we have
\begin{equation}
    \mathbb{E}(w^m_{\text{dynamic}})=w^m_{\text{static}}
\end{equation}

and
\begin{equation}
    r(w^m_{\text{dynamic}},\ell(f^m))\leq 0
\end{equation}
for all input modality $m$.           
\end{proof}

\section{Experimental details}
\label{appendix-B}
\subsection{Datasets details} 
$\circ$ \textbf{Senses recognition.} For NYUD-V2, following the standard split, we reorganize the 27 categories into 10 categories with 9 usual senes and one "others" category. For SUN RGB-D, following the previous work~\cite{han2021trusted}, we use the 19 major scene categories of SUN RGB-D, each of which contains at least 80 images.\\
$\circ$ \textbf{Image-text classification.} For FOOD-101, following the previous work~\cite{kiela2019supervised}, there are 60101 image-text pairs in the training set, 5000 image-text pairs in the validation set, and 21695 image-text pairs in the test set. For MVSA, we conduct the division strategy presents in~\cite{kiela2019supervised}. There are 1555 image-text pairs in the training set. The validation set contains 518 image-text pairs, and the test set contains 519 image-text pairs.

\subsection{Implementation details}
\textbf{Senses recognition.} For senses recognition task, we compare the proposed method with diverse multimodal fusion methods, including late fusion, align-based fusion, concatenated-based fusion and recent SOTA MMTM (attention-based fusion). Regarding late fusion, align-based fusion, concatenated-based fusion, we adopt the architecture of ResNet~\cite{he2016deep} pretrained on ImageNet~\cite{deng2009imagenet} as the backbone network for each modality. $\circ$ \textbf{Concatenate-based fusion} For concatenate-based fusion, we concatenate the representations extracted from different modalities by ResNet. Then a fully connection layer is deployed to map the multimodal representation to the target space. The dimensions of unimodal representation and common representation are 128 and 256 respectively. $\circ$ \textbf{Align-based fusion} The alignment fusion method is a re-implementation of~\cite{wang2016learning}. We deploy cosine distance to measure the similarity of representations. $\circ$ \textbf{MMTM} We follow the authors' implementation, where the squeeze ratio is set to 4. For all compared methods, we use Adam optimizer to train all above models with $\mathcal{L}_2 $ regularization and dropout~\cite{srivastava2014dropout}. The learning rate is 1e-4 and the dropout rate is 0.1. 

\textbf{Image-text classification.} For image-text classification, we compare the proposed method with diverse multimodal fusion methods, including late fusion, concatenated-bow fusion, concatenated-bert fusion and MMTM. For late fusion and concatenated-bert fusion, we adopt the architecutre of ResNet~\cite{he2016deep} pretrained on ImageNet~\cite{deng2009imagenet} as the backbone network for image modality and pre-trained Bert\cite{devlin2018bert} for text modality. For concatenated-Bow fusion, we use the Bow~\cite{pennington2014glove} to replace BERT for text modality. For the Bert models, we use BertAdam and regular Adam for the other models. The learning rate is 1e-4 with a warmup rate of 0.1. We adopt the early stop strategy based on validation accuracy.

For all above experiments, we conduct sampling during the whole training phase ($\rm T_s=1$). The hyperparameter $\lambda$ is set to $0.1$. Temperature parameters $\{\mathcal{T}^m\}^M_{m=1}$ are set to $1$.

\section{Additional results}
\subsection{Full results with  standard deviation}
\label{sec:app-C.1}
In this section, we present the full results with standard deviation in Tab.~\ref{tab:appendix-classification-1}, and Tab.~\ref{tab:appendix-ablation}.

\subsection{Different type of noise}
\label{sec:app-C.2}
We provide more results with different type of noise (i.e., salt-pepper noise with varying noise rate $\epsilon$) in Tab.~\ref{tab:appendix-classification-salt}. The results validate that the proposed method can improve the performance of multimodal fusion methods under different type of noise.

\begin{table*}[!htbp]
\vspace{-0.0cm}
\begin{center}
\caption{Full ablation study on NYU Depth V2. \label{ablation-appendix}}
\label{tab:appendix-ablation}
\center
{
\setlength{\tabcolsep}{1.9mm}
\begin{tabular}{cc|cccccccc}
\toprule
\multirow{1}{*}{\text{UAW}}  & \multirow{1}{*}{\text{$\mathcal{L}_{\text{reg}}$}}  &
\multicolumn{1}{c}{$\mathbf{\epsilon=0.0}$} & \multicolumn{1}{c}{$\mathbf{\epsilon=5.0}$}& \multicolumn{1}{c}{$\mathbf{\epsilon=10.0}$}& \multicolumn{1}{c}{$\mathbf{\epsilon=20.0}$}
\\
\midrule
  
\xmark         &\xmark                                          &${69.14\pm0.69}$    &$68.35\pm0.82$
       &${59.62\pm1.17}$    &$53.98\pm1.08$
            \\
                                                                             \xmark  &\cmark             
                                                         &${69.68\pm0.39}$  &${67.74\pm0.40}$
       &${61.35\pm0.34}$   &${58.26\pm0.1.13}$
                                      \\  \cmark  &\xmark             
                                                         &${70.06\pm0.1.03}$  &$\pmb{69.11\pm0.82}$ 
                                                         &${61.59\pm0.0.72}$   &$57.49\pm1.41$
                                                          \\\midrule \cmark  & \cmark            
                                                         &$\pmb{70.09\pm0.38}$   &${68.81\pm0.62}$
                                                         &${\pmb{61.62\pm0.31}}$  &$\pmb{58.87\pm0.40}$
                                                                                           \\ \bottomrule
\end{tabular}}
\end{center}
\vspace{-0.4cm}
\end{table*}

\begin{table*}[!t]

\vspace{0.0cm}
\begin{center}
\caption{Full comparison results when 50\% of the modalities are corrupted with Gaussian noise.\label{tab:appendix-classification-1}}

\center
{
\renewcommand{\arraystretch}{1.25}
\begin{tabular}{c|ccccc}
\toprule
\multirow{1}{*}{\text{Dataset}}  & \multirow{1}{*}{\text{Dynamic}} & \multirow{1}{*}{\text{Method}}    & \multicolumn{1}{c}{\text{$\mathbf{\epsilon=0.0}$}}& \multicolumn{1}{c}{\text{$\mathbf{\epsilon=5.0}$}}& \multicolumn{1}{c}{\text{$\mathbf{\epsilon=10.0}$}}
\\

\midrule
 \multirow{8}{*} {\text{\shortstack{NYU \\Depth V2}}} 
&\xmark &  \text{RGB} &${62.65\pm1.22}$  &${50.95\pm3.38}$ &${44.13\pm3.80}$ \\
&\xmark &  \text{Depth}  &${63.30\pm0.48}$   &${53.12\pm1.52}$ &${45.46\pm2.07}$ \\ 
&\xmark & {\text{Late fusion}} &${69.14\pm0.67}$  &${59.63\pm2.44}$ &${51.99\pm3.11}$ \\ 
&\xmark  & {\text{Concat}} &${70.31\pm0.80}$  &${59.97\pm2.14}$ &${53.20\pm3.50}$ \\ 
&\xmark  & {\text{Align}} &${70.31\pm1.28}$  &${59.47\pm1.84}$ &${51.74\pm3.41}$ \\ 
&\cmark  & {\text{MMTM}}  &${71.04\pm0.41}$  &${60.37\pm2.61}$ &${52.28\pm3.77}$ \\  
&\cmark  & {\text{TMC}}   &$\pmb{71.06\pm0.76}$  &${61.04\pm1.66}$ &${53.36\pm2.76}$  \\
\cline{2-6} &\cmark  & {\text{Ours}} &${70.09\pm0.97}$  &$\pmb{61.62\pm1.84}$ &$\pmb{55.60\pm2.09}$  \\

\midrule  \multirow{8}{*} {\text{\shortstack{SUN\\RGB-D}}} 
&\xmark &  \text{RGB} &${52.99\pm0.88}$  &${37.81\pm1.14}$ &${33.07\pm1.81}$   \\
&\xmark &  \text{Depth} &${56.78\pm0.19}$ &${48.40\pm1.11}$ &${42.94\pm1.63}$  \\ 
&\xmark  & {\text{Late fusion}} &${62.00\pm0.15}$  &${52.52\pm0.67}$ &${47.48\pm1.40}$ \\ &\xmark  & {\text{Concat}}&$\pmb{62.48\pm0.50}$  &${53.30\pm0.39}$ &${48.01\pm0.96}$ \\ 
&\xmark  & {\text{Align}} &${61.12\pm0.61}$  &${50.05\pm1.59}$ &${44.19\pm2.18}$ \\ &\cmark  & {\text{MMTM}} &${61.72\pm0.67}$  &${51.86\pm1.14}$ &${46.03\pm1.47}$  \\  &\cmark  & {\text{TMC}}  &${60.68\pm0.24}$  &${51.24\pm0.96}$ &${45.66\pm2.06}$ \\
\cline{2-6} 
&\cmark & {\text{Ours}} &${62.09\pm0.56}$  &$\pmb{53.40\pm0.89}$ &$\pmb{48.58\pm0.82}$ \\

\midrule  \multirow{8}{*} {\text{\shortstack{UMPC\\FOOD101}}} 
&\xmark &  \text{Bow}  &${82.50\pm0.18}$  &${61.68\pm0.71}$ &${41.95\pm0.54}$   \\
&\xmark &  \text{Img}  &${64.62\pm0.40}$  &${34.72\pm0.53}$ &${33.03\pm0.37}$   \\ 
&\xmark &  \text{Bert}  &${86.46\pm0.05}$  &${67.38\pm0.19}$ &${43.88\pm0.32}$   \\ 
&\xmark  & {\text{Late fusion}} &${90.69\pm0.12}$ &${68.49\pm3.37}$ &${57.99\pm1.59}$  \\ 
&\xmark  & {\text{Concatbow}}  &${70.77\pm0.09}$  &${38.28\pm0.26}$ &${35.68\pm0.69}$   \\ 
&\xmark  & {\text{Concatbert}} &${88.20\pm0.34}$  &${61.10\pm2.02}$ &${49.86\pm2.05}$   \\ 
&\cmark  & {\text{MMBT}} &${91.52\pm0.10}$  &${72.32\pm0.34}$ &${56.75\pm0.33}$   \\ 
&\cmark  & {\text{TMC}} &\textbf{${89.86\pm0.07}$}  &${73.93\pm0.34}$ &${61.37\pm0.21}$   \\ 
\cline{2-6} &\cmark  & {\text{Ours}} &$\pmb{92.92\pm0.11}$  &$\pmb{76.03\pm0.70}$ &$\pmb{62.21\pm0.25}$   \\

\midrule  \multirow{8}{*} {\text{\shortstack{MVSA}}} 
&\xmark &  \text{Bow}  &${48.79\pm7.05}$  &${42.20\pm6.40}$ &${41.57\pm6.28}$   \\
&\xmark &  \text{Img}  &${64.12\pm1.23}$  &${49.36\pm2.02}$ &${45.00\pm2.63}$   \\
&\xmark &  \text{Bert}  &${75.61\pm0.53}$  &${69.50\pm1.50}$ &${47.41\pm0.79}$   \\ 
&\xmark  & {\text{Late fusion}} &${76.88\pm1.30}$ &${63.46\pm3.46}$ &${55.16\pm3.60}$  \\ 
&\xmark  & {\text{ConcatBow}}  &${64.08\pm1.54}$  &${49.95\pm2.29}$ &${45.39\pm3.03}$   \\ 
&\xmark  & {\text{ConcatBert}} &${65.59\pm1.33}$  &${50.70\pm2.65}$ &${46.12\pm2.44}$   \\ 
&\cmark  & {\text{MMBT}} &${78.50\pm0.40}$  &${71.99\pm1.04}$ &${55.34\pm2.84}$   \\ 
&\cmark  & {\text{TMC}} &${74.87\pm2.24}$  &${66.72\pm4.55}$ &${60.35\pm2.79}$   \\ 
\cline{2-6} &\cmark  & {\text{Ours}} &$\pmb{78.07\pm1.10}$ &$\pmb{73.85\pm1.42}$ &$\pmb{61.28\pm2.12}$

\\ \bottomrule
\end{tabular}}
\end{center}
\vspace{0mm}
\end{table*}

\begin{table*}[!t]

\vspace{0.0cm}
\begin{center}
\caption{Full comparison results when 50\% of the modalities are corrupted with Salt-pepper noise.\label{tab:appendix-classification-salt}}

\center
{
\renewcommand{\arraystretch}{1.25}
\begin{tabular}{c|ccccc}
\toprule
\multirow{1}{*}{\text{Dataset}}  & \multirow{1}{*}{\text{Dynamic}} & \multirow{1}{*}{\text{Method}}    & \multicolumn{1}{c}{\text{$\mathbf{\epsilon=0.0}$}}& \multicolumn{1}{c}{\text{$\mathbf{\epsilon=5.0}$}}& \multicolumn{1}{c}{\text{$\mathbf{\epsilon=10.0}$}}
\\

\midrule
  \multirow{8}{*} {\text{\shortstack{NYU \\Depth V2}}} 
&\xmark &  \text{RGB} &${62.61\pm1.21}$  &${49.14\pm1.40}$ &${34.76\pm1.59}$ \\
&\xmark &  \text{Depth}  &${63.32\pm0.50}$   &${50.99\pm1.41}$ &${38.56\pm2.16}$ \\ 
&\xmark & {\text{Late fusion}} &${69.16\pm0.68}$  &${56.27\pm2.40}$ &${41.22\pm2.78}$ \\ 
&\xmark  & {\text{Concat}} &${70.44\pm0.89}$  &${57.98\pm2.08}$ &${44.51\pm2.91}$ \\ 
&\xmark  & {\text{Align}} &${70.31\pm1.28}$  &${57.54\pm2.50}$ &${43.01\pm2.66}$ \\ 
&\cmark  & {\text{MMTM}}  &$\pmb{71.04\pm0.41}$  &$\pmb{59.45\pm1.38}$ &${44.59\pm2.49}$ \\  
&\cmark  & {\text{TMC}}   &${71.01\pm0.75}$  &${59.34\pm1.03}$ &${44.65\pm2.30}$  \\
\cline{2-6} 
&\cmark  & {\text{Ours}} &${70.06\pm0.81}$  &${58.50\pm2.05}$ &$\pmb{45.69\pm2.79}$ \\ 
       
\midrule  \multirow{8}{*} {\text{\shortstack{SUN\\RGB-D}}} 
&\xmark &  \text{RGB} &${52.63\pm0.89}$  &${40.42\pm0.99}$ &${28.15\pm1.00}$ \\
&\xmark &  \text{Depth}  &${56.81\pm0.57}$   &${46.36\pm0.82}$ &${35.66\pm1.44}$ \\ 
&\xmark & {\text{Late fusion}} &${61.79\pm0.57}$  &${51.54\pm2.12}$ &${39.35\pm2.89}$ \\ 
&\xmark  & {\text{Concat}} &$\pmb{62.06\pm0.53}$  &${51.09\pm1.91}$ &${38.61\pm3.07}$ \\ 
&\xmark  & {\text{Align}} &${61.02\pm0.54}$  &${50.45\pm0.82}$ &${38.70\pm1.46}$ \\ 
&\cmark  & {\text{MMTM}}  &${61.80\pm0.40}$  &${51.09\pm0.77}$ &${38.38\pm1.56}$ \\  
&\cmark  & {\text{TMC}}   &${61.02\pm0.39}$  &${50.88\pm1.28}$ &${39.61\pm2.30}$  \\
\cline{2-6} &\cmark  & {\text{Ours}} &${61.89\pm0.49}$  &$\pmb{52.49\pm1.81}$ &$\pmb{40.53\pm2.79}$ \\  
       
\midrule  \multirow{8}{*} {\text{\shortstack{UMPC\\FOOD101}}} 
&\xmark &  \text{Bow}  &${82.43\pm0.18}$  &${60.83\pm0.54}$ &${41.56\pm0.33}$   \\
&\xmark &  \text{Img}  &${64.53\pm0.47}$  &${50.75\pm0.44}$ &${36.83\pm0.92}$   \\ 
&\xmark &  \text{Bert}  &${86.44\pm0.02}$  &${67.41\pm0.20}$ &${43.89\pm0.33}$   \\ 
&\xmark  & {\text{Late fusion}} &${90.66\pm0.16}$ &${77.99\pm0.54}$ &${58.75\pm0.99}$  \\ 
&\xmark  & {\text{Concatbow}}  &${70.68\pm0.12}$  &${55.01\pm0.33}$ &${38.81\pm0.62}$   \\ 
&\xmark  & {\text{Concatbert}} &${88.22\pm0.36}$  &${72.49\pm0.75}$ &${52.10\pm0.97}$   \\ 
&\cmark  & {\text{MMBT}} &${91.51\pm0.10}$  &${76.27\pm0.22}$ &${54.98\pm0.55}$   \\ 
&\cmark  & {\text{TMC}} &${89.86\pm0.07}$  &${77.86\pm0.41}$ &${60.22\pm0.43}$   \\ 
\cline{2-6} &\cmark  & {\text{Ours}} &$\pmb{92.90\pm0.13}$  &$\pmb{80.87\pm0.40}$ &$\pmb{61.60\pm0.20}$   \\

\midrule  \multirow{8}{*} {\text{\shortstack{MVSA}}} 
&\xmark &  \text{Bow}  &${48.82\pm7.08}$  &${42.23\pm6.43}$ &${41.60\pm6.31}$   \\
&\xmark &  \text{Img}  &${64.12\pm1.23}$  &${56.72\pm1.92}$ &${50.71\pm3.20}$   \\
&\xmark &  \text{Bert}  &${75.61\pm0.53}$  &${69.50\pm1.50}$ &${47.41\pm0.79}$   \\ 
&\xmark  & {\text{Late fusion}} &${76.88\pm1.30}$ &${67.88\pm1.87}$ &${55.43\pm1.94}$  \\ 
&\xmark  & {\text{ConcatBow}}  &${64.08\pm1.54}$  &${56.66\pm1.73}$ &${49.35\pm2.44}$   \\ 
&\xmark  & {\text{ConcatBert}} &${65.59\pm1.33}$  &${58.69\pm2.25}$ &${51.16\pm2.99}$   \\ 
&\cmark  & {\text{MMBT}} &$\pmb{78.50\pm0.40}$  &$\pmb{74.07\pm1.12}$ &${51.26\pm5.65}$   \\ 
&\cmark  & {\text{TMC}} &${74.87\pm2.24}$  &${68.02\pm3.07}$ &${56.62\pm3.67}$   \\ 
\cline{2-6} 
&\cmark  & {\text{Ours}} &${78.07\pm1.10}$ &${73.90\pm1.89}$ &$\pmb{60.41\pm2.63}$

\\ \bottomrule
\end{tabular}}
\end{center}
\vspace{0mm}
\end{table*}

\end{document}